\documentclass{article}
\usepackage{float}
\usepackage{amsmath, amsfonts}
 \usepackage[hidelinks]{hyperref}
\usepackage[utf8]{inputenc}
\usepackage[T1]{fontenc}
\usepackage[small]{caption}
\usepackage{tikz}
\usetikzlibrary{positioning,shapes,shadows,arrows,calc}
\tikzstyle{line}=[draw, very thick, black]
\tikzstyle{auxline}=[draw, very thin, black]
\tikzstyle{arrow}=[draw, -latex, very thick, black]
\tikzstyle{dot}=[draw, very thick, -latex , black, dashed]
\tikzstyle{ldot}=[draw, very thick, black, dashed]
\usetikzlibrary{fit}
\usepackage{subcaption}
\usepackage{todonotes}
\usepackage{times}
\usepackage{makecell}
\usepackage{verbatimbox}
\usepackage{booktabs}

\begin{document}
 
\bibliographystyle{plain}

\title{Autoencoders and Generative Adversarial Networks for Imbalanced Sequence Classification}
\author{Stephanie Ger$^1$, Yegna Subramanian Jambunath$^2$, Diego Klabjan $^3$}
\date{$^1$ Department of Engineering Sciences and Applied Mathematics, Northwestern University, Evanston, Illinois, USA \\
$^2$ Center for Deep Learning, Northwestern University, Evanston, Illinois, USA\\
$^3$ Department of Industrial Engineering and Management Sciences, Northwestern University, Evanston, Illinois, USA\\
\today}
%

\maketitle

\begin{abstract}
Generative Adversarial Networks (GANs) have been used in many different applications to generate realistic synthetic data. We introduce a novel GAN with Autoencoder (GAN-AE) architecture to generate synthetic samples for variable length, multi-feature sequence datasets as existing GAN models cannot generate synthetic data and associated labels. In this model, we develop a GAN architecture with an additional autoencoder component, where recurrent neural networks (RNNs) are used for each component of the model in order to generate synthetic data to improve classification accuracy for a highly imbalanced medical device dataset. In addition to the medical device dataset, we also evaluate the GAN-AE performance on two additional datasets and demonstrate the application of GAN-AE to a sequence-to-sequence task where both synthetic sequence inputs and sequence outputs must be generated. To evaluate the quality of the synthetic data, we train encoder-decoder models both with and without the synthetic data and compare the classification model performance. We show that a model trained with GAN-AE generated synthetic data outperforms models trained with synthetic data generated both with standard oversampling techniques such as SMOTE and Autoencoders as well as with state of the art GAN-based models.
\end{abstract}
	
	Dealing with imbalanced datasets is the crux of many real world classification problems. These problems deal with complex multivariate data such as variable length, multi-feature sequence data. Canonical examples can be found in the finance world, for example, questions related to stock market data of several securities or credit card fraud detection often deal with sequence data with many features. Other imbalanced data problems include questions in the medical field such as tumor detection and post surgery prognosis \cite{thoracic}. In each of these problems, false positives are more desirable than false negatives, they require sequential data, and the classes are imbalanced. 

	Class imbalances in datasets oftentimes lead to increased difficulty in classification problems as many machine learning algorithms assume that the dataset is balanced. There are two general approaches to improve classification accuracy for unbalanced datasets. One method is algorithmic, for example, a modified loss function can be used so that misclassifications of minority labeled data are penalized more heavily than misclassifications of majority labeled data \cite{costCNN}. The other is to decrease data imbalances in the training set either by ensembling the data or by generating synthetic training data to augment the amount of data in the minority set. 
	
	This motivates the development of methods to improve classification accuracy on variable length, multi-feature sequence data. Given a sequence of $T$ feature vectors, we want to predict labels of length $S$ for the sequence. Oftentimes it is not obvious how to apply methods for unbalanced data to sequence data in a way that takes advantage of the fact that sequential events have the potential to be highly correlated. SMOTE \cite{SMOTE} is widely used for oversampling, but does not capture the sequential dimension. Enhanced Structure Preserved Oversampling (ESPO) \cite{ESPO} allows one to generate synthetic data that preserves the sequence structure, however it requires that the feature vector has only a single feature at each of the $T$ time points and that the output label is a scalar. As there is no obvious extension to the case where there are multiple features at each time point and the output is also a sequence of labels, the situations where ESPO can be applied are limited. 

	We develop a method based on deep learning models for sequences in order to decrease data imbalances of sequence data with an arbitrary number of features. We call each feature vector, $x_i \in R^n$, an event in the sequence. We consider the use of generative adversarial networks (GANs) to generate synthetic data. Here, we build a generative model that generates both the feature vectors in a sequence as well as the corresponding labels. We benchmark this synthetic data generation technique against a number of models. We demonstrate that the model trained on the GAN with Autoencoder based synthetic data outperforms the baseline model and other standard synthetic data generation techniques. For each of the synthetic data generation methods, we train a sequence-to-sequence model \cite{sutskever2014sequence} on the dataset that outputs a sequence with the same length as the label sequence. In addition to benchmarking against existing synthetic data generation techniques, we also train a model on the unaugmented dataset. All of the models are embedded within the standard ensemble approach. On all of our datasets, we observe that the GAN with Autoencoder based synthetic data generation model significantly improves over the non-GAN baseline models by 15\% to 127\% depending on the dataset. We also benchmark the GAN with Autoencoder model against a model trained on RGAN generated synthetic data \cite{RGAN}. Like other existing GAN-based models for generating synthetic data, the RGAN model is unable to generate both synthetic data and labels that the GAN with Autencoder model is capable of generating. We observe that GAN with Autoencoder synthetic data outperforms models trained on RGAN synthetic data and can be applied to a wider range of datasets.	
The main contributions are as follows:
	\begin{enumerate}
	\item a novel synthetic data generation technique that uses a GAN with an Autoencoder component to generate synthetic data for variable length, multi-feature sequential data in a way that preserves the structure of sequences for both feature vectors and labels;
	\item a computational study of existing imbalanced classification techniques on highly imbalanced sequential datasets. 
	\end{enumerate}
In the next section, we discuss relevant literature. Section 3 discusses all of the models, while the computational results are presented in Section 4. 

\section{Literature Review}
	Many methods exist for imbalanced data. The majority of these methods are developed for non-sequential data and generally take one of two approaches. The first approach is algorithmic and either involves altering the loss function or performance metric in a way that emphasizes the correct classification of the minority set. The second approach is to decrease the data imbalance either by resampling or by generating synthetic minority data such that the training data is more balanced .
	
	The benefit of using algorithmic methods is that they have a straightforward application to sequence data as we can calculate the loss and accuracy the same way for both a vector and a scalar. Methods that are commonly used include a weighted loss function in which the loss of misclassifying minority data is greater than the loss of misclassifying majority data \cite{imb-cost,costCNN}. We implement a weighted loss function in all our models.  
	
	In contrast to the algorithmic methods, we can instead consider data level methods that strive to balance the two classes. There have been many different methods that are developed to balance the dataset without generating synthetic minority data. Since these methods alter how the training set is built, applying them to sequence data is straightforward. Both ensembling and data sampling techniques fall under this category. Ensemble methods take the original training set and build subsets of the training set such that the sizes of the minority and majority sets are more balanced \cite{ensembles}. On the other hand, other methods for dataset creation involve over- or under-sampling \cite{kubat1997addressing}. Ensemble methods generally outperform over- and under-sampling methods alone so we use ensembles in all our experiments. 
	
	Another data level method that can mitigate the class imbalance problem is to generate synthetic minority data. SMOTE \cite{SMOTE} is one of the most widely used methods for generating synthetic minority data. For this method, synthetic data is generated via interpolation between nearest neighbors in the minority set. There are many extensions to SMOTE that aim to increase classification performance by sharpening the boundary between the two classes. One such example is ADASYN \cite{ADASYN}, which explores the composition of the nearest neighbors to determine how many synthetic data points to generate and how to generate them. Neither SMOTE nor ADASYN cannot be used to oversample sequence data because these methods build a synthetic feature vector by independently interpolating between the real data points, so the framework cannot capture correlation in time. While models that use an autoencoder and apply SMOTE in the latent space have been developed to oversample sequence data, these models do not consider how to oversample sequence labels for training sequence-to-sequence models \cite{DOPING}.
	
	Structure Preserving Oversampling and ESPO are methods that exist for dealing with unbalanced sequence data \cite{SPO,ESPO}. To generate synthetic sequence data, these methods use the covariance structure of the minority data in order to build synthetic minority data that captures the sequential structure. They are developed for single feature sequences and there is not a straightforward extension to data that has multiple features for each event. This is because we cannot calculate the covariance matrix for each feature independently since features may interact with each other in different ways at different events. \textcolor{black}{The only non-GAN based model that exists for dealing with unbalanced multivariate sequence data with variable label sequence length is a kernel based oversampling method \cite{CRJKer}. In this model, a recurrent model is used to generate a representation of the input sequence which are used to compute a kernel. Oversampling and classification is then done in the kernel feature space. As this model can be applied to variable length, multivariate sequence data, we benchmark our proposed model against this method.}
	
	Another method for synthetic data generation are GANs \cite{GAN}. This model pits a generator model, which generates synthetic data, and a discriminator model, which tries to distinguish between real and synthetic data, against each other. By pitting the models against each other, it trains both the generator and discriminator, and once the generator has been trained, we can use it to generate synthetic minority data. While this approach has been applied to oversample both image data \cite{zenati2018efficient,guo2019discriminative,douzas2018effective} and sequence data with models such as SeqGAN and RGAN \cite{yu2017seqgan,RGAN}, these models cannot generate synthetic input sequences and corresponding labels as they are designed to be trained only on data with a single label. Therefore, if there are multiple minority labels, it is not clear how these models can be applied. GAN based models designed for sequence data have been used for synthetic text generation, but as this architecture is not designed for classification, the sequence class is not considered, which suggests that the models cannot generate both a sequence and the associated labels. \textcolor{black}{GAN based models have been used to build imbalanced sequence classification models, but the benefit of generating GAN-based synthetic minority data is that it allows for flexibility during classification model selection. \cite{rezaei2018multi}.} We benchmark our proposed model against a model trained with RGAN generated synthetic data for the single label classification tasks. 
	
	Both SMOTE and GAN based synthetic data generation techniques have been shown to improve classification performance for certain types of highly imbalanced datasets such as image data or single feature sequences. \textcolor{black}{While methods such as RGAN have been developed for multivariate sequence generation, they are not designed to simultaneously generate input sequences and sequence labels. So while these oversampling methods may improve a classifier's performance, unlike other data-level and algorithmic methods such as weighted loss functions or oversampling minority data, they have not yet been developed and applied to generic sequence data.}  

	
\section{Baseline Approaches} 
	We assume that we have sequences $x = (x_1, \dots x_T) \in \mathcal{X}$ and associated labels $y = (y_1, \dots, y_L) \in \mathcal{Y}$ where each $x_i$ has $n$ features and $L$ labels to predict. \textcolor{black}{Each of the labels $y_{\ell}$ for $\ell \in L$ is a class label, either 0 or 1. We consider binary labels at each prediction step, but multi-class labels can be considered as well.} Sequence length $T$ can vary by  sequence. We also assume there is a dominant label sequence called majority and all other label sequences are minority. Since we focus on minority sequences, all our synthetic oversampling methods also work with no modification in the presence of multiple majority classes. For the baseline model, we consider a sequence-to-sequence (seq2seq) architecture. This is an encoder-decoder architecture where the entire sequence is represented by an $s$ dimensional hidden vector $h^0_T$, the encoder hidden state at the final event. We then use this vector, $h^0_T$, as the input to the decoder model at each event. The model can be written as
	\begin{align*}
	h^0_t &= f^0_{\theta_E}(h^0_{t-1},x_t), \text{ } t \in [1,T]\\
	h^1_{\ell} &= f^1_{\theta_D}(h^1_{\ell-1}, h^0_T), \text{ } \ell\in [1, L]\\
	o_{\ell} &= \text{softmax}(h^1_{\ell})
	\end{align*}
where $f^0_{\theta_E}, f^1_{\theta_D}$ are cell functions such as LSTM or GRU and $o_\ell$ is the $\ell^{th}$ predicted label \cite{sutskever2014sequence}. In our experiments, we use a seq2seq model with attention \cite{attention} and a weighted loss function where the weights are proportional to class balance as the classification method. The output of this seq2seq model is the same length as the label sequence. We ensemble the data into $K$ ensembles where each ensemble contains a subset of the majority data and all of the minority training data and in inference, we average the predictions from each ensemble. \textcolor{black}{In order to evaluate the synthetic data generation techniques, we train seq2seq models without synthetic minority data as a baseline and compare the classification results for seq2seq models trained with and without synthetic data.}

\subsection{SMOTE and ADASYN for Sequential Data}

	In a straightforward application of SMOTE to sequences, we reshape $x$ to a vector and then apply the SMOTE algorithm directly to $x$. In addition, by reshaping the label $y$, we can interpolate between the label vectors associated with the samples used to generate the synthetic sample. As it does not make sense to interpolate between variable length inputs, we compare the straightforward SMOTE application on the datasets where sequences are all of the same length. 
	
	\textcolor{black}{As SMOTE cannot be applied to variable length inputs, we consider ADASYN to generate synthetic variable length inputs.} We first train an autoencoder on minority data. Using the trained autoencoder on the minority data, we obtain $h^0_T \in \mathbb{R}^s$ for each sequence. Once we have embedded the sequence, we can then run the ADASYN algorithm to get $\hat{h}^0_T$. We use ADASYN instead of SMOTE as there is only a single interpolation weight for ADASYN, so it is clear how to interpolate between the label vectors. Next, we can use the decoder half of the autoencoder to lift $\hat{h}^0_T$ back to $\hat{x}$. The benefit of this approach is that the encoded minority data captures the structure of the sequence. We use the ADASYN interpolation weight, $w^i$, for each synthetic sample to get the label vector, $\hat{y}$, associated with $\hat{x}$ using the equation \[\hat{y} = y_i + w^i(y_i - y_j)\] \textcolor{black}{for sequence prediction classification tasks}. 

\section{Generative Adversarial Network with Autoencoder Based Synthetic Data}\label{GAN-AE}

	We develop a GAN that is capable of generating both sequences, $x$, and associated label vectors $y$. As in any GAN model, we must build both a generator and a discriminator and train the models by pitting them against each other. The model that we discuss is based on the improved Wasserstein GAN (IWGAN) \cite{iwgan,wgan}. This model differs from the RGAN model as the RGAN model does not include an autoencoder component, but instead consists of an LSTM based generator and discriminator. As the RGAN does not employ a sequence-to-sequence architecture in the generator, it cannot be used to generate both sequential data and labels the way that GAN-AE can. A model trained on a dataset augmented with GAN-AE generated synthetic data is benchmarked against a model trained on an unaugmented dataset, a dataset augmented with ADASYN generated data, and a dataset augmented with RGAN generated data to determine how this proposed model compares to both well established synthetic data techniques as well as more recent GAN-based synthetic data techniques.   
	
	For the generator model, $G_{\phi_{EN_1}, \phi_{EN_2}}(z,x,y)$ we use a seq2seq model with LSTM cells to get hidden state sequences $h_x$ and $h_y$. We include an additional argument $z$ to initialize the cell state for the generator. For the true data, we set $z$ to 0 and for the fake data we use $z \sim \mathcal{N}(0,I)$. The model is able to distinguish between $x$ and $y$ since $x$ is the input for the generator encoder and $y$ is the input for the generator decoder. The parameters $\phi_{EN_1}$ and $\phi_{EN_2}$ correspond to $x$ and $y$, respectively. 
 The discriminator model, $D_{\phi_{D_1},\phi_{D_2}}(h_x, h_y)$ uses a seq2seq model trained on the hidden sequences $h_x$ and $h_y$ to get a real valued output, $c$.  As in the generator, $\phi_{D_1}$ are parameters corresponding to $x$ and $\phi_{D_2}$ to $y$. The loss function compares the outputs from the discriminator model for the real and fake data. 
				
\begin{figure}[h]
\centering
\scalebox{0.75}{\begin{tikzpicture}[
input/.style={rectangle, draw, very thick, minimum size=5mm},
nn/.style={rectangle, blue, draw, very thick, minimum size=5mm, fill = blue!20},
rnn/.style={circle, blue, draw, very thick, minimum size = 5mm,  fill = blue!20},
enc/.style={circle, blue, draw, very thick, minimum size = 5mm, fill = blue!60},
sum/.style = {circle, red, draw, very thick, minimum size = 5mm, fill = red!60, text=black},
every label/.style={align=center}]

\node[nn](enc1){$\phi_{EN_1}$};
\node(aux0)[right = 7.5mm of enc1]{};
\node(a0)[above = 5 mm of aux0]{};
\node[nn](enc2)[right = 5mm of a0]{$\phi_{EN_2}$};
\node[input](f)[left= of enc1, align=center]{Fake \\ $z$};
\node[input](x)[below = of enc1, align=center]{True \\ $x$};
\node[input](y)[right= 16.5mm of x, align=center]{True \\ $y$};
\node[draw, dashed, label = left:{Encoder \\ \\ \\ }, fit=(enc1) (enc2)]{};
\node[draw, dashed, label = left:{Generator }, fit=(x) (y) (f) (enc1) (enc2)]{};

\node[nn](dec1)[above= 25mm of enc1]{$\phi_{DE_1}$};
\node(aux1)[right = 7.5mm of dec1]{};
\node(a1)[above = 5mm of aux1]{};
\node[nn](dec2)[right = 5mm of a1]{$\phi_{DE_2}$};
\node[draw, dashed, label = left:{Decoder}, fit=(dec1) (dec2)]{};
\node[input](yh)[above = 10.75 mm of dec2]{$\hat{y}$};
\node[input](xh)[above = 18 mm of dec1]{$\hat{x}$};

\node[nn](dis1)[right = 30 mm of aux0]{$\phi_{D_1}$};
\node(aux2)[right = 7.5mm of dis1]{};
\node(a2)[above = 5mm of aux2]{};
\node[nn](dis2)[right= 5mm of a2]{$\phi_{D_2}$};
\node[draw, dashed, label = above:{Discriminator}, fit=(dis1) (dis2)]{};
\node[input](c)[right= 5mmof dis2]{$c$};

\node(a3)[right =6.5mm of enc2]{};
\node(a4)[below = 22.5mm of a3]{};
\path[ldot](enc2) -| (a4.center);
\node(a5)[right = 8 mm of a4]{};
\node(a6)[right = 21 mm of a5]{};
\path[ldot](a4.center) -- (a5.center); 
\path[dot](a5.center) -- node [right] {$h_x$} (dis1.south);
\path[ldot](a5.center) -- (a6.center);
\path[dot](a6.center) -- node [right, pos=0.325] {$h_y$} (dis2.south);

\path[line](enc1.east) -| (a0.west);
\path[arrow](a0.west) --(enc2.west);
\path[line](dec1.east) -| (a1.west);
\path[arrow](a1.west) --(dec2.west);
\path[line](dis1.east) -| (a2.west);
\path[arrow](a2.west) --(dis2.west);

\path[arrow](f) -- (enc1);
\path[arrow](x) -- (enc1); 
\path[arrow](y) -- (enc2);
\path[dot] (enc1) -- node [right, pos=0.65] {$h_x$}(dec1);
\path[dot] (enc2) -- node [right, pos=0.35] {$h_y$}(dec2);
\path[arrow](dec1) -- (xh);
\path[arrow](dec2) -- (yh);
\path[arrow](dis2) -- (c);

\end{tikzpicture}}

\caption{Overview of GAN model. Sequences and labels are used as input to GAN and both the discriminator and decoder use the outputs from the generator model.}
\label{fig:GAN} 
\end{figure}
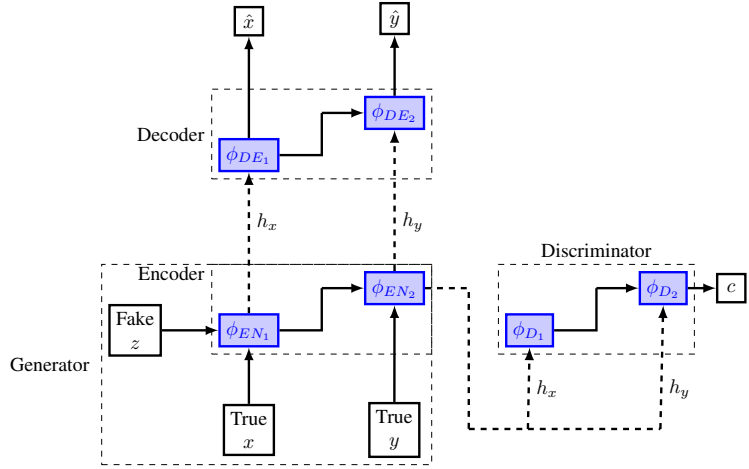

We also need a component of the model to lift $h_x$ and $h_y$ to $\hat{x}$ and $\hat{y}$ respectively. Therefore, we have a seq2seq based autoencoder, $A_{\phi_{EN_1},\phi_{EN_2},\phi_{DE_1},\phi_{DE_2}}(x,y)$, that takes as input $x$ and $y$, creates hidden sequences $h_x$ and $h_y$, and then reconstructs $\hat{x}$ and $\hat{y}$. The autoencoder shares the encoding part with the generator. This GAN architecture differs from existing GAN-based synthetic data generation methods as each of the three components of the GAN with Autoencoder model are comprised of LSTM encoder-decoder architectures in order to generate both minority sequences and associated labels.

In Figure~\ref{fig:GAN}, the GAN with autoencoder structure is sketched out. For model training, we use the loss function
	\begin{align}
	\mathcal{L} & = \mathbb{E}[D_{\phi_{D_1}, \phi_{D_2}}(G_{\phi_{EN_1},\phi_{EN_2}}(z,x,y))]  \nonumber \\
	&-\mathbb{E}[D_{\phi_{D_1},\phi_{D_2}}(G_{\phi_{EN_1}\phi_{EN_2}}(0,x,y))] \nonumber \\
	&+ \lambda\mathbb{E}\left[\left(\bigg|\bigg|\nabla D_{\phi_{D_1},\phi_{D_2}}(G_{\phi_{EN_1}\phi_{EN_2}}(0,x,y))\bigg|\bigg|_2 -1\right)^2\right] \nonumber \\
	&+ \mu \mathbb{E}\left[\bigg|\bigg|(x,y) -A_{\phi_{EN_1},\phi_{EN_2},\phi_{DE_1},\phi_{DE_2}}(x,y)\bigg|\bigg|_2^2\right] \label{eq1}
	\end{align}
	where $\lambda$ and $\mu$ are tunable hyperparameters. All expectations are with respect to the minority sequences $(x,y)$. \\
	
	 During training, we want to prevent the discriminator from learning too quickly so that the generator can learn. We use Adam \cite{adam}, and set the discriminator learning rate lower than the generator learning rate to prevent the discriminator from learning too quickly. \textcolor{black}{To further slow down discriminator training, we add noise based on $ \mathcal{N}(0,\sigma)$ to generator outputs with initial $\sigma$ set to 1 and decrement $\sigma$ during model training \cite{ganhacks}}. We set $\lambda$ approximately equal to 10 as discussed in the IWGAN paper and tune $\mu$ until we reach a ratio such that the discriminator does not learn too quickly and that the autoencoder loss decreases during training \cite{iwgan}. For our datasets, we find that $\mu \approx 0.01\lambda$ results in reasonable generator, discriminator and autoencoder loss curves.
	We train the generator, discriminator and autoencoder weights on different batches of data. We update the weights of each component of the model by considering the corresponding terms in the loss function. For example, the weights associated with the generator, $\phi_{EN_1}$ and $\phi_{EN_2}$, are present in all terms of the loss function so $\mathcal{L}$ is used to update the generator weights. The exception is the autoencoder as the weights of the encoder component of the autoencoder are shared with generator, so the weights $\phi_{EN_1}$ and $\phi_{EN_2}$ are not updated along with the rest of the autoencoder weights. For datasets with a single label prediction, we consider a GAN with autoencoder model, where instead of a seq2seq architecture for each of the model components, we use LSTM cells and the input to the generator is $x$ and $z$. We then assign the minority label to generated minority data. 
	
	Once we have trained the generator in conjunction with the discriminator and autoencoder, we can use the generator and the decoder part of the autoencoder to generate synthetic minority data. As this model is trained only on the minority dataset, we require a reasonably sized minority training set. In our experiments, we consider minority training sets with at least 1000 samples\textcolor{black}{, though it is possible to train a GAN for synthetic data generation on a smaller dataset}. We generate 3 synthetic samples from each minority sample in the training dataset by feeding in vectors $z \sim \mathcal{N}(0,1)$ into model and using the autoencoder output as synthetic minority data. \textcolor{black}{Synthetic samples are generated from the trained GAN weights at every 100 epochs and the weights that result in the highest validation F1-score on the seq2seq classification model are used.} We expect that with random noise $z$ will slightly perturb the minority data in order to generate novel synthetic minority samples instead of simply oversampling existing minority data. The GAN with Autoencoder model allows for the simultaneous generation of both the sequences and associated label vectors. This method should improve on the ADASYN with autoencoder model as it allows for the simultaneous generation of both the sequences and associated label vectors. 
	
\section{Computational Study}
We consider three imbalanced datasets\footnote{Code and data are available at \href{https://github.com/code-submission/imbalanced-sequence-classification}{\url{https://github.com/code-submission/imbalanced-sequence-classification}}}. Each of these datasets consists of multi-feature sequence data where the data imbalance is less than 5\% (it can be as low as 0.1\%). The first dataset is a proprietary medical device dataset where the data comes from medical device outputs. The second dataset we consider is a sentiment analysis dataset that classifies IMDB movie reviews as positive or negative \cite{imdb}. Though the data is initially balanced, for this paper, we downsample the positive class in order to use it for an anomaly detection task. Lastly, we consider a power consumption dataset\footnote{\href{https://archive.ics.uci.edu/ml/datasets/individual+household+electric+power+consumption}{Individual household electric power consumption Dataset} }where the goal is to predict if voltage consumption changes significantly. A class corresponds to whether the voltage change is considered significant. For the medical device dataset and IMDB sentiment dataset, we make a single label prediction and thus we consider the seq2one model for both these datasets. For the power consumption dataset, we consider both the seq2seq and seq2one tasks to show that the GAN with autoencoder (GAN-AE) generated synthetic data improves model performance in both cases.

For each of the datasets, the data is ensembled into 10 ensembles such that each ensemble contains all of the minority data and a random subset of the majority data. For models trained with synthetic data, we generate synthetic data for each of the 10 ensembles and train models on the augmented dataset. Sequences in each dataset are front-padded to the maximum sequence length for model training. The GAN based oversampling method is implemented using Tensorflow and the remaining models are implemented using Keras with Tensorflow. All models are trained with an adaptive optimizer and on a single GPU card. \textcolor{black}{The GAN with Autoencoder model discussed in Section \ref{GAN-AE} requires 1.6 GiB and between 30.71 and 64.48 gFlops to train depending on the dataset.}  For each dataset, we tune the number of layers and number of neurons of the baseline model. Models are trained with up to 5 layers and up to 150 neurons in each layer for all models regardless of synthetic data augmentation. The hyperparameters that return the highest validation F1-score are reported for each dataset and model. We use the best performing hyperparameter setting for a model trained without any synthetic data as the baseline model.

\textcolor{black}{We also benchmark our model against the kernel based oversampling model \cite{CRJKer}. Due to the high dimensionality of the multivariate temporal data for each of the datasets, we cannot implement the full model where both the cycle reservoir with jumps (CRJ) model and the kernel parameter are optimized. Therefore, we consider only the data-kernel model, where the kernel is learned on the input data, not a representation of the data. During training, we update kernel parameters on batches of the training data as the full dataset is too large to fit in memory. We embed the training data in the kernel feature space and oversample the training data. To compute test F1-scores, we select a random batch of training data to embed the test data in the kernel feature space. We then train an SVM model where parameters are determined using cross validation, and compute the performance metrics on the test data. For each dataset, we compare the GAN-AE based synthetic data model and the kernel-based oversampling model for a single run. For this model, we report test F1-scores of 0.005, 0.0 and 0.034, respectively for the medical device, IMDB, and power datasets respectively. We show that the LSTM baseline models and the proposed GAN-based synthetic data method both outperform the kernel based oversampling model. It is possible that batching the data makes training the kernel parameter difficult, leading to a poor kernel space feature representation, and subsequently the low test F1-scores across all datasets.}

	For each dataset, we report the F1 score for the baseline model without synthetic data, a model trained with the GAN-AE based synthetic minority data, \textcolor{black}{a model trained with minority data generated with RGAN}, and a model trained with minority data generated with an autoencoder and SMOTE in the latent space as a baseline for synthetic minority training data \cite{RGAN}. For each model, we report the minority class F1 score on the test set. If there are multiple minority classes, we report the average F1 score of the minority classes. In addition to reporting the F1-score, we also consider the G-mean and PR AUC metrics for both the baseline model and the model trained with GAN-AE synthetic data for datasets where the difference is statistically significant in order to get a complete picture of how the two models compare. We do not consider the additional metrics on the remaining models as they underperform the baseline model.

\subsection{Medical Device Data} In this dataset, the data is a sequence of readouts from medical devices and the labels indicate if a user error occurs. We predict if user error will occur within the next hour based on the sequence of actions recorded by the medical device. The sequence length is on average 50 and there are around 50 features. We have on order of 1 million samples and less than 1\% of the samples are from the minority class. We make 5 runs, each one with a different seed, and thus each run has different ensemble models.

Comparing the results of each of the proposed methods against the baseline in Table \ref{tab:med-comp}, we observe that the only method that significantly improves classification accuracy is the GAN-AE based synthetic data model with p-value = 0.01 based on the t-test. Surprisingly, using the ADASYN Autoencoder generated synthetic data leads to a substantial decrease in the F1-score, suggesting that this synthetic data technique does not capture the structure of the minority data. This suggests that interpolation in the autoencoder latent space is not sufficient, and the GAN component of the autoencoder is necessary. \textcolor{black}{We observe that the GAN-AE synthetic data model yields the highest F1-score across all seeds and that the average GAN-AE test F1-score is slightly higher than the average RGAN test F1-score. This suggests that the GAN-AE model performs about as well as the RGAN model and can be applied to a broader range of datasets.}

\begin{table}[H]
\begin{center}
\caption{Test F1-Scores for Each Seed}
\resizebox{0.65\textwidth}{!}{\begin{tabular}{c r r r r}
\toprule
Run & Baseline & \thead{GAN-AE \\ Synthetic Data} & \thead{ADASYN \\Autoencoder}  & \thead{RGAN \\Synthetic Data}\\\midrule
0 & 0.79\% & 2.02\% & 0.52\% & \textbf{2.04\%} \\\
1 & 1.77\% & \textbf{3.15\%} & 0.30\% & 1.92\%  \\
2 & 1.28\% & 2.06\% & 0.50\% & \textbf{2.24\%} \\
3 & 1.29\% & 1.85\% & 0.49\% & \textbf{2.44\%} \\
4 & 0.68\% & 1.79\% & 0.52\% & \textbf{2.13\%}\\ \midrule
Average & 1.16\% & \textbf{2.17\%} & 0.47\% & 2.15\%   \\ 
Standard Deviation & 0.44\% & 0.50\% &  \textbf{0.09\%} & 0.18\% \\
\bottomrule
\end{tabular}}
\label{tab:med-comp}
\end{center}
\end{table}

We see in Table~\ref{tab:med-metrics} that the model trained on the GAN-AE based synthetic data outperforms both the baseline model and the model trained on RGAN generated synthetic data on the G-mean and PR AUC metrics.

\begin{table}[H]
\begin{center}
\caption{Performance Metrics on Test Set Averaged Across Runs}
\resizebox{0.8\textwidth}{!}{\begin{tabular}{c r r r r r r }
\toprule
 &\multicolumn{2}{c}{Baseline} & \multicolumn{2}{c}{GAN-AE}  & \multicolumn{2}{c}{RGAN}\\
\midrule
 & G-mean & PR AUC & G-mean & PR AUC & G-mean & PR AUC\\ \midrule
Average & 19.8\% & 0.0027 & \textbf{24.1\%}&\textbf{0.0031}  & 18.52 & 0.0029 \\
Standard Deviation &3.18\%& 0.0002 & \textbf{0.04\%}& 0.0002 & 2.8\% & \textbf{0.0001}\\
\bottomrule
\end{tabular}}
\label{tab:med-metrics}
\end{center}
\end{table}

%

To explore how the models trained on the synthetic data improve on the baseline models, we examine the difference between the confusion matrix of predictions on the test set for a model trained with and without the GAN-AE based synthetic data. In Table~\ref{tab:medical-device-conf-comp}, we note that a number of false negatives and false positives in the baseline model are converted to true positives and true negatives, respectively in the model trained on the GAN-AE based synthetic data. That is, the improvement in classification accuracy of the model trained with the GAN-AE based synthetic data is due to a decrease in both false negatives and false positives. 

\begin{table}[ht]
\begin{center}
\caption{Differences Between Predictions for GAN-AE Minority and Baseline Models}
\resizebox{0.4\textwidth}{!}{\begin{tabular}{c r r }
\toprule
& True Majority & True Minority \\
\midrule
Predicted Majority & 60 & -1 \\
Predicted Minority & -60 & 1 \\
\bottomrule
\end{tabular}}
\label{tab:medical-device-conf-comp}
\end{center}
\end{table}

Examining the classification of true minority and synthetic minority samples in the GAN-AE based synthetic data training set, we observe that the trained model is better at correctly classifying the synthetic minority samples than the true minority samples which is interesting. For run 0, the F1-score for the true minority training samples is 0.4036 while the F1-score for the synthetic minority training samples is 1. This also reveals that the model overfits since the test F1-score is much lower. This is not surprising for such a heavily imbalanced dataset.

\subsection{Sentiment} We consider all reviews under 600 words long and front pad reviews so that all samples in our dataset are of length 600. We then use the GoogleNews trained word2vec model to embed the dataset. In order to make this dataset imbalanced, we downsample the positive reviews to create two datasets where the positive reviews comprise 1\% and 5\%  of the training set respectively and then ensemble the training dataset. The resulting dataset is comprised of around 25 thousand samples with 20\% in test. Training models on this dataset is computationally expensive because of the sequence length, so we only consider a single run for these experiments. 


\begin{table}[H]
\begin{center}
\caption{Test F1-Scores}
\resizebox{0.6\textwidth}{!}{\begin{tabular}{c r r r r}
\toprule
Data Imbalance & Baseline & \thead{GAN-based\\ Synthetic Data} & \thead{ADASYN \\Autoencoder} &\thead{RGAN \\Synthetic Data} \\\midrule
1\%& 7.80\% & \textbf{17.76\%} & 0.00\% & 12.70\% \\
5\% & \textbf{56.75}\% & 52.85\% & 9.47\% & N/A \\ \bottomrule
\end{tabular}}

\label{tab:sentiment-results}
\end{center}
\end{table}

In Table~\ref{tab:sentiment-results}, we compare the results of each of the proposed methods against the baseline. \textcolor{black}{While the RGAN and GAN-AE models both improve the F1-score over the baseline, we observe that the F1-score for the GAN-AE model is 5\% higher than the RGAN F1-score.} With 5\% imbalance, the baseline model performance on the ensembles is high enough that the anomaly detection methods we consider do not improve performance. We consider data imbalances from 1\% to 10\%, but as the accuracy is high for the model trained on the 5\% imbalance dataset without synthetic data, we do not report the accuracies on the 10\% imbalance dataset. This suggests that both the GAN-AE synthetic data generation technique and existing models are only effective for highly imbalanced datasets. \textcolor{black}{We do not train a model with the RGAN generated synthetic data on the 5\% imbalance dataset as we have determined the imbalance is not high enough to observe improved classification performance with synthetic data. }

\begin{table}[H]
\begin{center}
\caption{Performance Metrics on Test Set}
\resizebox{0.6\textwidth}{!}{\begin{tabular}{r r r r r r}
\toprule
\multicolumn{2}{c}{Baseline} & \multicolumn{2}{c}{GAN-AE} & \multicolumn{2}{c}{RGAN} \\
\midrule
G-mean & PR AUC & G-mean & PR AUC &  G-mean & PR AUC  \\ \midrule
22.7\% & 0.031 & \textbf{56.4\%}&\textbf{0.062} & 29.31\% & 0.031 \\
\bottomrule
\end{tabular}}

\label{tab:imdb-metrics}
\end{center}
\end{table}

We see in Table~\ref{tab:imdb-metrics} that the model trained on the GAN-based synthetic data outperforms the baseline model and the RGAN model on both the G-mean and PR AUC metrics for the 1\% imbalance. We do not consider the G-mean or PR AUC metrics for the dataset with 5\% imbalance as the GAN-based synthetic data does not improve classification accuracy for that level of data imbalance.  

%

\begin{figure}[h]
\begin{center}
\caption{t-SNE Embedding of Minority Training Data}
\begin{subfigure}[t]{.5\textwidth}
\centering
\includegraphics[width=\textwidth]{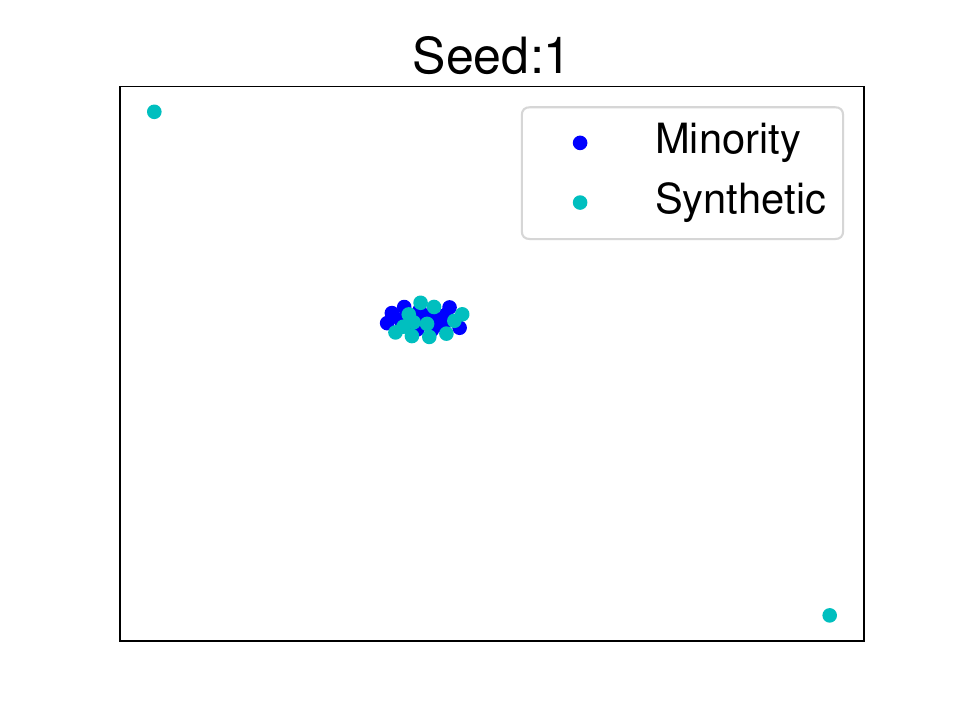}
\label{fig:sentiment-tsne-0}
\end{subfigure}%
\begin{subfigure}[t]{.5\textwidth} 
\centering
\includegraphics[width=\textwidth]{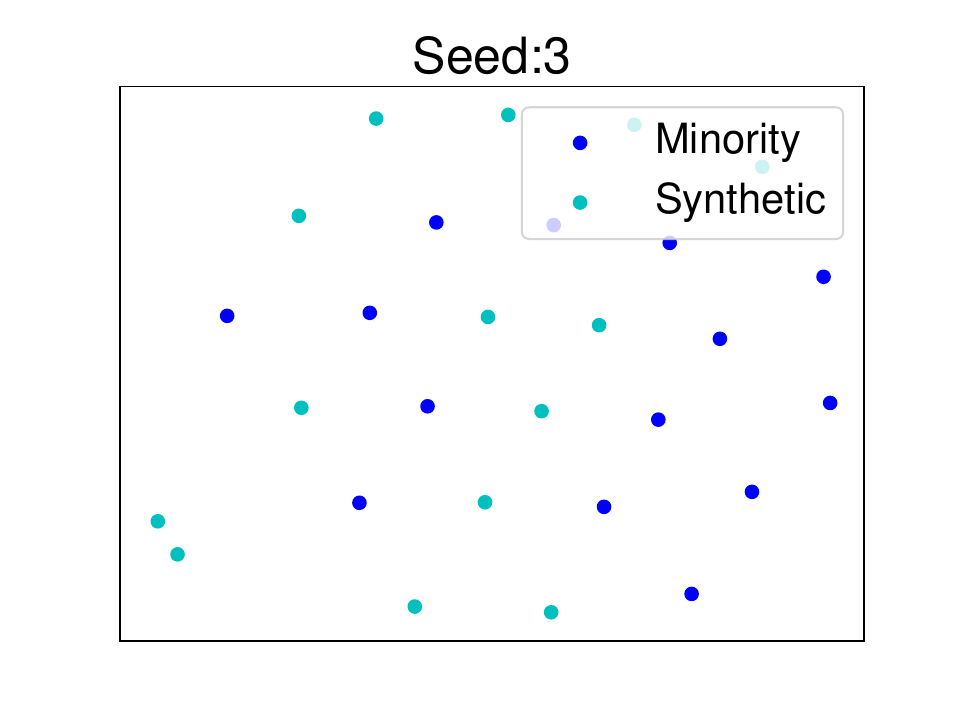}
\label{fig:sentiment-tsne-2}
\end{subfigure}

\label{fig:sentiment-tsne}
\end{center}
\end{figure}

For this dataset, we conclude that 5\% imbalance is an upper bound for which the proposed anomaly detection techniques can be used. However, studying classification of true minority and synthetic minority samples in the GAN-AE based synthetic data, we notice that the trained model correctly identifies all minority samples in the training set, both true and synthetic. This suggests that the sentiment analysis task is an easier task.

To understand how well the GAN-based synthetic data training set is able to capture the structure of the minority data, we use t-SNE to embed a subset of the true and synthetic minority training data so it can be visualized. In Figure~\ref{fig:sentiment-tsne}, it is clear that the true minority data forms a cluster and all but two synthetic minority samples belong to this cluster. T-SNE embeddings can vary from run to run, so we include embeddings initialized with two different random seeds to show that there is consistency between runs. As the synthetic samples are part of this cluster, it suggests that for the most part, the synthetic minority data successfully mimics the minority data. 

\subsection{Power} We use a dataset of power usage in a given household in trying to predict if voltage usage changes significantly. Sequences are of length 20 and there are 6 features. We have around 2 million sample and approximately 2\% of the samples are in the minority class. As this dataset is not padded, we compare our GAN-AE based synthetic data technique against a model trained with SMOTE generated synthetic data. 

\begin{figure}[h]
\centering
\caption{Bar Plot of Test F1-Scores for Each Model with Confidence Intervals for Models Trained on Multiple Datasets}
\includegraphics[scale=0.28]{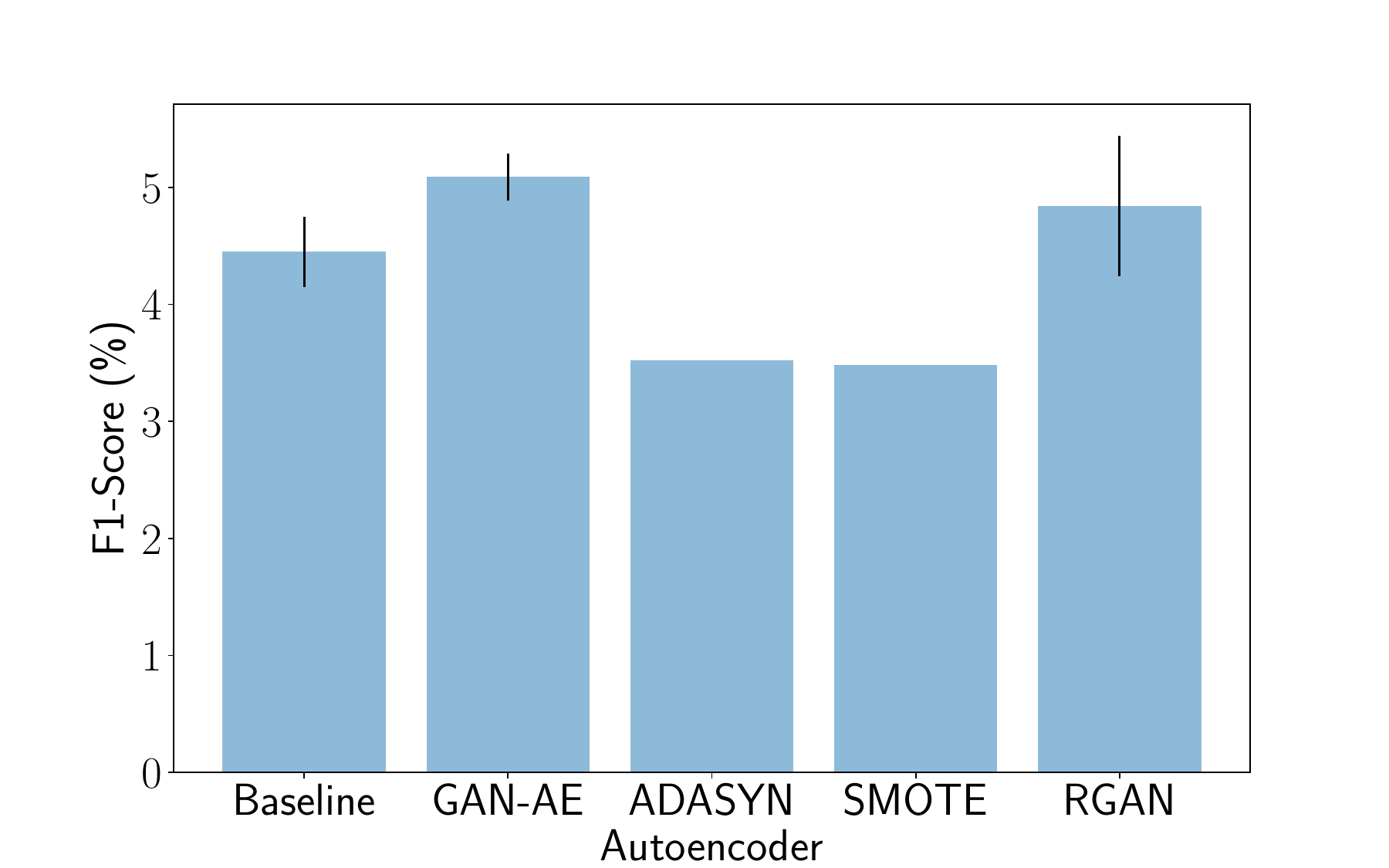}

\label{fig:power-results}	
\end{figure}

Comparing the results of each of the proposed methods against the baseline in Figure \ref{fig:power-results}, we conclude that the only method that significantly improves the F1-score is the model trained on the GAN-AE based synthetic data. To test the significance of this improvement, we generate ensembles using 5 different seeds and train a baseline and GAN-AE based synthetic data model on each run. In the five runs, the average baseline F1-score is 4.45\%, the average F1-score for the GAN-AE based synthetic data is 5.10\%, and the improvement with the GAN-AE based synthetic data is significant with p-value=0.023 based on the t-test. \textcolor{black}{The average F1-score for the models trained with the RGAN synthetic data is 4.84\%, but the improvement over the baseline model is not significant as the p-value is 0.154. As the ADASYN and SMOTE models return substantially lower F1-scores than the seq2seq model trained without additional synthetic data, we do not conduct multiple runs for these models.}

Note that the relative difference in the F1-score between the baseline model and the GAN-AE based synthetic data model is about $15\%$ and lower than either the Medical Device or Sentiment dataset. As the Power dataset has fewer features than the other two datasets, we observe that the GAN-AE based synthetic data is better able to capture the data structure for more complex sequences. 

\begin{table}[t!]
\begin{center}
\caption{Performance Metrics on Test Set Averaged Across Runs}
\resizebox{0.7\textwidth}{!}{\begin{tabular}{c r r r r r r}
\toprule
 &\multicolumn{2}{c}{Baseline} & \multicolumn{2}{c}{GAN-AE} &  \multicolumn{2}{c}{RGAN}\\
\midrule
 & G-mean & PR AUC & G-mean & PR AUC & G-mean & PR AUC \\ \midrule
Average & 16.3\% & 0.02 & \textbf{20.9\%}& 0.02 &  17.52\% & 0.02\\
Standard Deviation &\textbf{ 1.3\%}& 0 &1.6\%& 0 & 1.5\% & 0 \\
\bottomrule
\end{tabular}}

\label{tab:power-metrics}
\end{center}
\end{table}

We see in Table~\ref{tab:power-metrics} that the model trained on the GAN-AE based synthetic data outperforms the baseline model and the model trained with the RGAN synthetic data on the G-mean metric, but not the PR AUC metric. It is interesting that there is a much larger disparity in both the G-mean and F1-scores between the baseline model and the model trained with GAN-AE based synthetic data, yet the PR AUC scores are identical.

%

On this dataset, we also consider sequences where the associated label vectors are of length 4 by predicting if the voltage change is significant for 4 time periods. As before, sequences are of length 20. We consider a sample as minority if the voltage change is significant in any of the 4 time periods. Approximately 7\% of the data is in the minority class. We only consider the GAN-AE based synthetic data model on this dataset as it is the only model that improves on the baseline in Figure~\ref{fig:power-results}. The average baseline F1-score is 0.25\% and the average F1-score for the GAN-based synthetic data is 0.59\%. Though the imbalance is lower, it is unsurprising that the F1-score is so low as we are making 4 predictions for each sequence. We do not do multiple runs for this dataset as the relative F1-score increase is high. We conclude that the GAN-AE based synthetic data can be used to improve model performance for datasets with label sequences. 

\section{Conclusions}	
We have presented several techniques for synthetic oversampling in anomaly detection for multi-feature sequence datasets. Models were evaluated on three datasets where it was observed that GAN-AE based synthetic data generation outperforms all non-GAN models on all datasets\textcolor{black}{, including a model designed to oversample imbalanced multivariate temporal data}. \textcolor{black}{On the three datasets, the GAN-AE model slightly outperforms the RGAN model, which is incapable of generating synthetic seq2seq minority data.} We also note that GAN-based synthetic data yielded larger classification F1-score increases over other models for datasets with more features. Furthermore, we provide evidence that the GAN-based synthetic data is capable of capturing the structure of minority data. We also demonstrate that GAN-based synthetic data generation techniques can be applied to datasets with label sequences. Finally, we provide evidence that synthetic oversampling is beneficial for datasets with substantial imbalances (less than 5\% in our datasets).

\nocite{9565320}
\nocite{9197818}
\bibliography{main}

\end{document}